\documentclass{article}

    \usepackage[preprint]{neurips_2020}

\usepackage[utf8]{inputenc} 
\usepackage[T1]{fontenc}    
\usepackage{hyperref}       
\usepackage{url}            
\usepackage{booktabs}       
\usepackage{amsfonts}       
\usepackage{nicefrac}       
\usepackage{microtype}      
\usepackage{xcolor}         
\usepackage{graphicx}
\usepackage{subfigure}
\usepackage{wrapfig}
\usepackage{tikz}

\usepackage{amsmath,amssymb} 
\usepackage{mathtools}
\usepackage{multirow}
\usepackage{kotex}
\usepackage{subfigure}
\usepackage{amsfonts}
\usepackage{breqn}
\usepackage[ruled,vlined]{algorithm2e}
\SetKwComment{Comment}{$\triangleright$}{}

\usepackage{mathtools}

\newcommand{\etal}{\textit{et al}. }
\newcommand{\ie}{\textit{i}.\textit{e}., }
\newcommand{\eg}{\textit{e}.\textit{g}., } 
\newcommand{\et}{\textit{et al.}}
\newcommand{\mathbbm}[1]{\text{\usefont{U}{bbm}{m}{n}#1}}

\title{Loss-based Sequential Learning  \\ 
for Active Domain Adaptation}

\makeatletter
\newcommand{\printfnsymbol}[1]{%
  \textsuperscript{\@fnsymbol{#1}}%
}
\makeatother

\author{%
     \textbf {Kyeongtak Han} \\
  Inha University \\
  \texttt{han00127@inha.edu} \\
  \and
   \textbf{Youngeun Kim} \\
  Yale University \\
   \texttt{youngeun.kim@yale.edu} \\
  \and
   \textbf{Dongyoon Han} \\
  NAVER AI Lab \\
  \texttt{dongyoon.han@navercorp.com} \\
  \and
   \textbf{Sungeun Hong}\\
  Inha University \\
  \texttt{csehong@inha.ac.kr} \\
}

\begin{document}

\maketitle

\begin{abstract}
Active domain adaptation (ADA) studies have mainly addressed query selection while following existing domain adaptation strategies. However, we argue that it is critical to consider not only query selection criteria but also domain adaptation strategies designed for ADA scenarios. This paper introduces sequential learning considering both domain type (source/target) or labelness (labeled/unlabeled). We first train our model only on labeled target samples obtained by loss-based query selection. When loss-based query selection is applied under domain shift, unuseful high-loss samples gradually increase, and the labeled-sample diversity becomes low. To solve these, we fully utilize pseudo labels of the unlabeled target domain by leveraging loss prediction. We further encourage pseudo labels to have low self-entropy and diverse class distributions. Our model significantly outperforms previous methods as well as baseline models in various benchmark datasets.
\end{abstract}

\section{Introduction}
\label{introduction}

Although unsupervised domain adaptation (UDA) has shown promising results across various fields \cite{wang2018deep,zhao2020review}, the performance gap between UDA and its supervised setting, \ie usage of target labels, is significant. 
The non-negligible performance gap is the main obstacle to applying UDA to real-world problems.
Recently, semi-supervised domain adaptation has been actively investigated to alleviate such challenges.
In the semi-supervised settings, target samples to be labeled are randomly selected \cite{li2021learning} or sampled at an even rate across all classes \cite{saito2019semi,jiang2020bidirectional}.
However, random labeling is less efficient than sampling based on specific criteria when limited annotation budget.
Additionally, applying uniform sampling across classes is impractical because the target samples' labels are presumably unknown.
For this reason, active domain adaptation (ADA) that aims to annotate the most informative target samples under domain shift automatically has emerged \cite{su2020active,fu2021transferable}.

One of the main challenges in the ADA task is how to select the most informative target samples under domain shift with limited annotation budgets.
Early works \cite{su2020active,huang2018cost} use uncertainty \cite{joshi2009multi} or diversity \cite{sener2018active} as sample selection criteria.
However, uncertainty estimation on the target domain is usually miscalibrated, resulting in sampling outliers or redundant instances as indicated by \cite{Prabhu_2021_ICCV}. 
Furthermore, domain similarity used as diversity cannot guarantee discriminative feature space under domain shift \cite{fu2021transferable}.
Therefore, more recent work \cite{Prabhu_2021_ICCV} jointly uses uncertainty and diversity as query selection criteria. 
Rangwani \et \cite{Rangwani_2021_ICCV} adopt submodular subset selection, which is a combination of uncertainty, diversity, and representativeness score of samples. Ma \et \cite{ma2021active} primarily utilizes uncertainty and diversity score. The state-of-the-art TQS \cite{fu2021transferable} adopts query by committee scheme \cite{beluch2018power} with uncertainty and domain similarity, which utilizes consensus between multiple predictions.

Critically, most previous ADA methods have focused primarily on query selection, but have not investigated how to deal with the labeled source, labeled target, and unlabeled target domains after query selection.
Several methods \cite{su2020active, fu2021transferable} perform ADA by integrating the newly labeled target domain and the existing labeled source domain into a single data pool.
Prabhu \et \cite{Prabhu_2021_ICCV} do not directly integrate the labeled source and labeled target domains, but treat them as one dataset in the training process.
We argue that it is critical to consider not only how to choose informative unlabeled target samples, but also how to treat labeled target samples, considering domain shift.

\begin{figure}[t]
	\begin{center}
		\def\arraystretch{0.5}
		\begin{tabular}{@{}c@{}c}
			\includegraphics[width=0.36\linewidth]{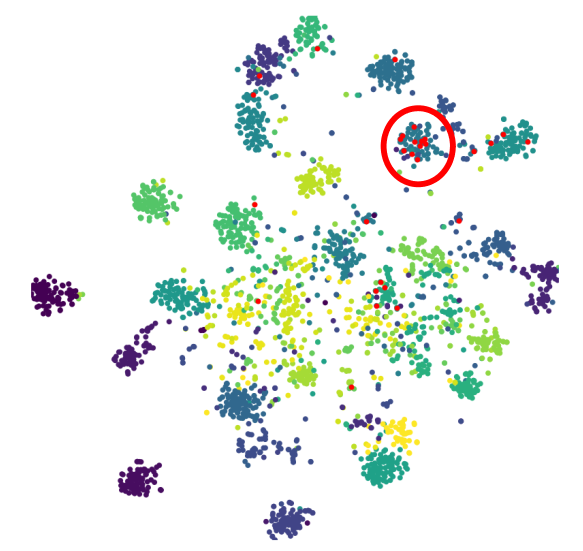} \vspace{1mm} & \qquad \qquad 
			\includegraphics[width=0.36\linewidth]{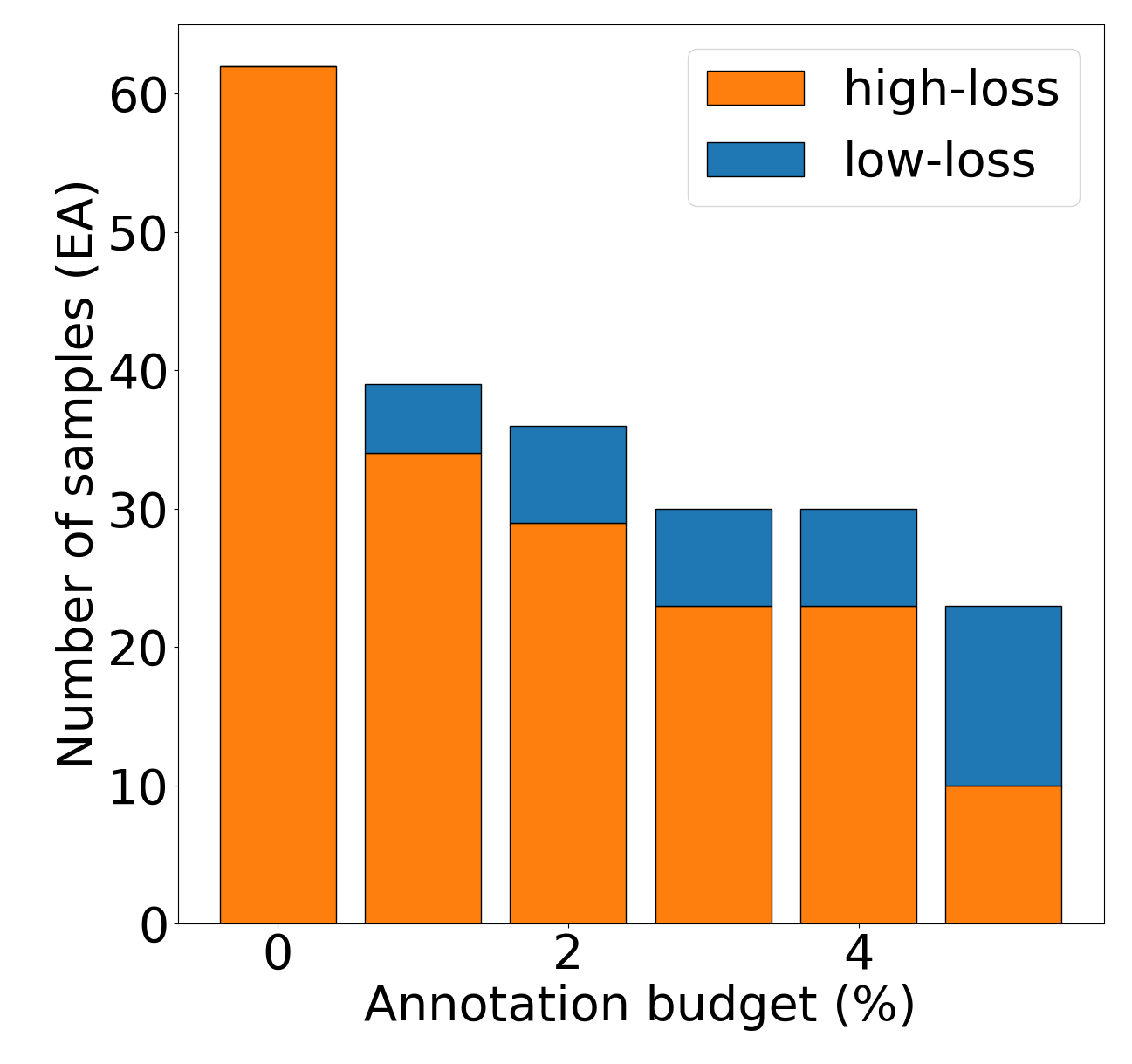}  \\
			{(a) Low diversity} & \qquad \qquad  {(b) Misprediction rate}
		\end{tabular}
	\end{center}
	\vspace{-2mm}
	\caption{Limitations of loss-based query selection in active domain adaptation. 
		(a) Labeled samples with high-loss (indicated in red) show low diversity in the target domain. Detailed description can be found in Section \ref{ssec:diversity_issue}.
		(b) Mispredicted high-loss samples gradually decrease while the low-loss samples increase in unlabeled target domain, which indicates unuseful samples are labeled.
	}
	\label{fig:loss_figures}
\end{figure}

\begin{table}[]
	\caption{Comparison of the proposed method on various benchmark datasets, in which $\Delta$ represents the difference in accuracy (\%) between `Learning loss + DA' and `Ours'. }
	\label{table:prob_def1}
	\vspace{-3mm}
	\vskip 0.15in
	\small
	\centering
	\tabcolsep=0.19cm
	\begin{tabular}{@{}lcccc@{}}
		\toprule
		& Baseline & Learning loss \cite{yoo2019learning} + DA  & Ours   & \quad   $ \Delta$ \\ \toprule
		Office-31   & 80.0     & 83.6               & 92.2 &  \quad  +8.6     \\
		Office-Home & 58.3     & 68.1               & 75.6 &  \quad   +7.5       \\
		VisDA       & 44.7     & 85.8               & 86.8 &  \quad +1.0      \\
		\bottomrule
	\end{tabular}
	\vspace{-3mm}
\end{table}

In this paper, we propose sequential ADA learning, which takes into account domain type (source/target) or labelness (labeled/unlabeled).
As a query selection criterion, we exploit \cite{yoo2019learning} that actively selects unlabeled samples with high loss predicted by an auxiliary module. 
However, high-loss samples selected by loss prediction often do not help model training and cannot handle the diversity issue, as shown in Fig. \ref{fig:loss_figures}.
We analyzed the model's misprediction rate by dividing unlabeled samples into two groups (high or low) according to their expected loss values.
Surprisingly, as the annotation budget increases, the number of falsely predicted high-loss samples decreases; while the number of low-loss samples increases.
Furthermore, loss-based query selection does not account for sample diversity, which is exacerbated under domain shift.

To solve the aforementioned issues, we introduce a sequential adaptation strategy by learning with a small number of labelness samples and then using a large number of unlabeled samples aggressively.
We first train the model using only the labeled target samples given by the oracle.
Labeled target samples contribute to increasing discriminative target representation but occupy only a small fraction of the target domain.
Therefore, training a model based on a small number of labeled target samples is difficult to reflect the overall target distribution.
To address this, we fully utilize a number of unlabeled target domains via pseudo labeling.
We further encourage pseudo labels to have low self-entropy and diverse class distributions to increase the reliability of the target pseudo labels.

Our contributions are as follows. (i) We propose a novel loss-based ADA learning that sequentially utilizes a small number of ground-truth labels by oracle and numerous pseudo labels obtained by elaborated self-learning. (ii) We analyze the limitations of loss-based query selection that have not been used in active domain adaptation, and present a simple but effective solution that outperforms the baseline models as shown in Table \ref{table:prob_def1}. (iii) The proposed method is superior to the existing ADA methods by a large gap in various datasets. 

\section{Related Work}
\subsection{Active learning}
Over the past decades, a number of selection criteria for active learning (AL) have been suggested and these can be divided into three categories: uncertainty, committee, and diversity.
Uncertainty-based approaches using conﬁdence \cite{wang2016cost}, entropy \cite{huang2018cost}, or best-vs-second-best \cite{joshi2012scalable} have been widely used because of its simplicity and high-computational efficiency.
However, most uncertainty-based approaches use task-specific uncertainty measures.
Query-by-committee \cite{beluch2018power} that leverages consensus among an ensemble of multiple classiﬁers can handle this issue, allowing a wide range of applications.
Diversity-based selection \cite{nguyen2004active,sener2018active} is also task-agnostic because selecting the samples that represent the overall distribution of the unlabeled data pool does not depend on particular tasks. 

Overall, most of the existing approaches require task-specific architectures or are computationally inefficient, especially for the recent deep networks.
Motivated by the fact that deep networks are trained by minimizing loss regardless of the number of tasks, task types, and model complexity, \cite{yoo2019learning} propose a new task-agnostic approach.
Considering the model's loss value as uncertainty, they attach a loss prediction module to the main network and trained this module to estimate the loss of unlabeled samples.
Critically, since all the criteria previously used in active learning are designed considering a single domain, they are not non-transferable. Therefore, they could not select the most informative sample for annotation under domain shift.

\subsection{Domain adaptation}
Unsupervised domain adaptation (UDA) has been investigated a lot as it can transfer knowledge of a labeled source domain to an unlabeled target domain \cite{gong2012geodesic}.
Even though accessing all label information of the target domain requires expensive labeling costs, labeling only a few samples and applying them to the UDA process is cost-effective.
For this reason, semi-supervised domain adaptation (SSDA) methods have been proposed across various areas \cite{saito2019semi,jiang2020bidirectional,li2021learning}.
Active domain adaptation (ADA) \cite{chattopadhyay2013joint} is similar to SSDA \cite{saito2019semi,jiang2020bidirectional} in that both can access few labeled target data.
While SSDA approaches label target samples randomly or according to predetermined rules, ADA models automatically select target samples to be labeled.

Previous methods mainly utilize uncertainty and diversity as sample selection criteria. 
Su \etal \cite{su2020active} use diversity from importance weights as well as uncertainty via entropy for query selection.
Prabhu \etal \cite{Prabhu_2021_ICCV} identifies target samples that are uncertain and representative in feature space and the training scheme is based on the mini-max entropy (MME) \cite{saito2019semi}.
Rangwani \etal \cite{Rangwani_2021_ICCV} combine uncertainty, diversity and representativeness of samples via submodular subset and train the labeled source, labeled target and unlabeled target samples with improved VADA \cite{shu2018dirt}.
Query by committee scheme was also used for ADA and training only labeled source and labeled target, which has shown state-of-the-art performance \cite{fu2021transferable}.
Critically, constructing multiple classiﬁers and their respective data pipelines requires high computational cost.
Instead of focusing on query selection criteria, we propose a new sequential process that leverages loss-based query selection.

\section{Method}

\begin{figure*}[]
	\centering
	\includegraphics[width=\textwidth]{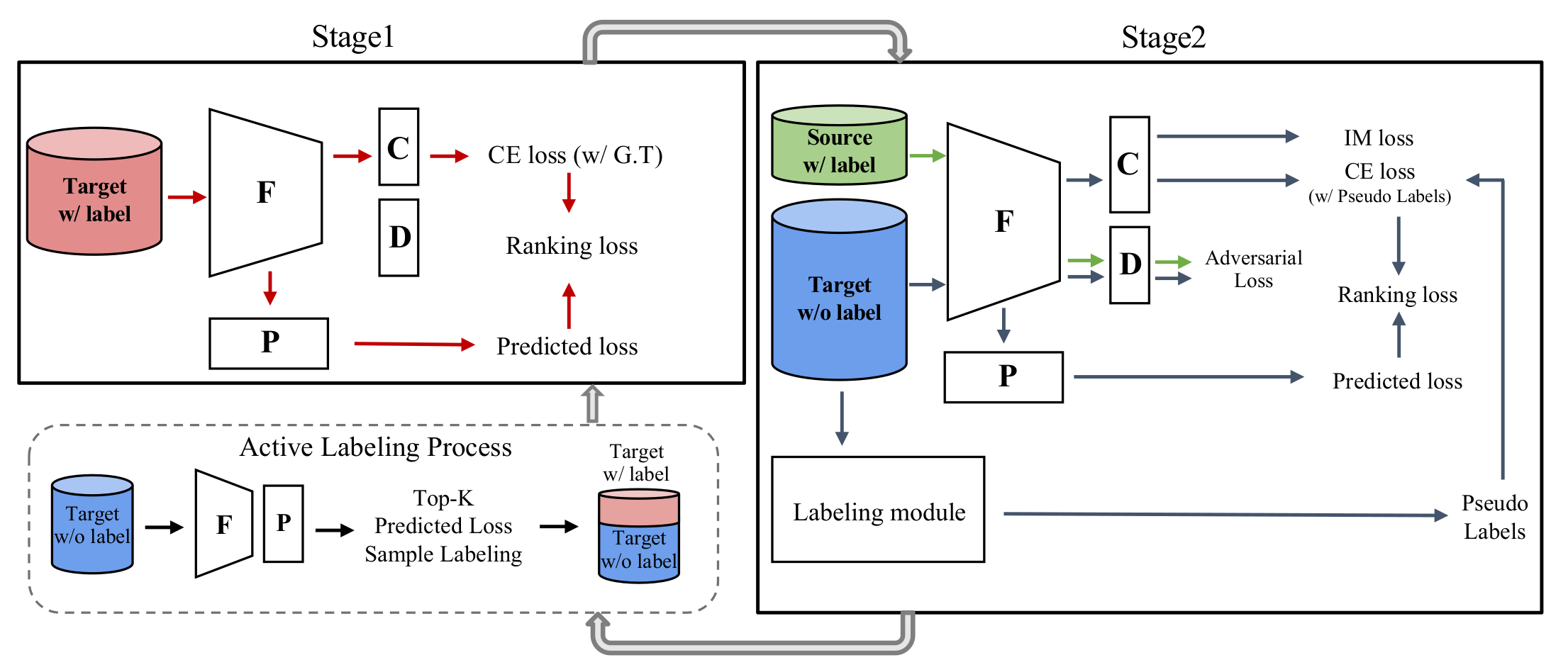}
	\caption{
		Outline of the proposed ADA framework considering domain type (source/target) and labelness (labeled/unlabeled).
	}
	\label{fig:archtiecture}
	\vspace{-2mm}
\end{figure*}

\vspace{-2mm}
\subsection{Overall framework}
In active domain adaptation (ADA) scenarios, we are given labeled source domain $\mathcal{X}_{s} = \{(x^i_{s}, y^i_{s})\}^{N_{s}}_{i=1}$ and unlabeled target domain  $\mathcal{X}_{t_u} = \{(x^i_{t_u})\}^{N_{t_u}}_{i=1}$ where $N_{s}$ and $N_{t_u}$ refer to the total number of the labeled source and unlabeled target domains, respectively.
As model training progresses, a part of the target domain is labeled by oracle according to query selection and denoted as $\mathcal{X}_{t} = \{(x^i_{t}, y^i_{t})\}^{N_{t}}_{i=1}$. 
Once the labeled target domain $\mathcal{X}_{t}$ is obtained, the model is trained in a semi-supervised manner with the existing $\mathcal{X}_{s}$ and  remaining $\mathcal{X}_{t_u} \gets \mathcal{X}_{t_u} \setminus \mathcal{X}_{t}$. 
Such query selection and semi-supervised domain adaptation are repeated until the annotation budget $B$ is reached.

The main goal of our framework is to alleviate domain discrepancies between the fully labeled source domain and the partially-labeled target domain by leveraging query selection.
Fig. \ref{fig:archtiecture} illustrates the outline of the proposed ADA framework.
The main difference between the proposed method and individual-related work is the learning schemes rather than the designs of each module or architecture.
Unlike most existing ADA methods \cite{fu2021transferable,rangwani2021s3vaada}, we separate model training for labeled source domain and labeled target domain as in \cite{liang2020we,tzeng2017adversarial}.
Concretely, we pre-train a feature extractor $F$ and a classifier $C$ with the labeled source domain $\mathcal{X}_{s}$.
We then freeze the classifier and alternately perform domain adaptation and query selection within the annotation budget $B$ to optimize each sub-network.

As a first step, we train the model mainly on labeled target samples $\mathcal{X}_{t}$ obtained by oracle, similar to existing ADA approaches.
We exploit loss-based query selection in our framework.
However, if the model is trained only on labeled target samples by loss-based query selection, the model could be overfitted by a small number of labeled samples with low diversity.
To increase generalization power in the target domain, we fully utilize a majority of unlabeled target samples by leveraging pseudo labeling and information maximization loss \cite{krause2010discriminative,shi2012information}.
Simultaneously, we train an adversarial domain discriminator to reduce domain discrepancy. 
After that, the samples with the top-$K$ high-loss predicted by the auxiliary loss predictor $P$ in the unlabeled target domain pool $\mathcal{X}_{t_u}$ are labeled as described in Algorithm \ref{alg:Query_algorithm}.

\begin{algorithm}[]
	\SetAlgoLined
	\textbf{Input}: unlabeled target domain: $\mathcal{X}_{t_u}$, non-trainable modules : ($F$, $P$), annotation budget: $B$  \\
	\textbf{Output}: set of queries $Q(X)$\\
	  initialize a predicted loss set L = []  \\
	 initialize an active sample set Q = []   \\
	\For{$i=1$ to ${N_{t_u}}$}{
		 $\hat{l}^i$ = $P(F(x_{t_u}^i))$\\
		  L $\gets$ L $\cup$ $\hat{l}^i$ 
	}
	\For{$i=1$ to $B$}{
		 Q $\gets$ Q $\cup$  ${\mathrm{argmax}_{x_{t_u}^i}}$(L)\\
		  L $\gets$ L $\setminus$ $\hat{l}^i$
	}
	\caption{Query selection}
	\label{alg:Query_algorithm}
\end{algorithm}

\vspace{-2mm}
\subsection{Model training with labeled target domain}
An intuition of loss-based query selection is to actively label high-loss samples because samples with high loss have a high probability of being incorrectly recognized by the model.
To this end, we train the auxiliary loss predictor $P$ using labeled target samples $\mathcal{X}_{t}$ obtained through query selection previously.
Specifically, once labeled target domain $\mathcal{X}_{t} = \{(x^i_{t}, y^i_{t})\}^{N_{t}}_{i=1}$ is obtained by oracle, we train the feature extractor $F$ using conventional cross-entropy loss $\mathcal{L}_{ce}$. 
The obtained actual loss value $\mathcal{L}_{ce}$ is then used as a ground-truth label for the auxiliary loss predictor $P$.
The scale of the real loss $\mathcal{L}_{ce}$ steadily decreases as the model training progresses, so we use margin ranking loss as follows:  
\vspace{-2mm}
\begin{equation}
\begin{split}
L_{loss} &= \mathbb{E}_{x_t \in \mathcal{X}_{t}}{max(0, -\mathbbm{1}({l_{n},l_{m}) \cdot ({\hat{l}_n-\hat{l}_m}) + \Delta}})  \\
& s.t.\text{  }\mathbbm{1}({l_{n},l_{m}}) = \begin{cases}+1,& \text{if }\text{ } l_{n} > l_{m} \\-1,&\text{otherwise},\end{cases}
\end{split}
\label{eq:predLoss}
\vspace{-5mm}
\end{equation}

where the $n$ and $m$ refer to the ranking pair index in the training mini-batch and  $\Delta$ is a pre-defined margin.
Unlike the conventional loss-based query selection for active learning \cite{yoo2019learning}, the gradient from our loss predictor $P$ flows to the feature extractor $F$, so they are learned jointly. 
This approach is effective for generating pseudo labels of unlabeled target samples by making our feature extractor $F$ learn loss under domain shift.

\vspace{-2mm}
\subsection{Sequential adaptation with unlabeled target domain}
After training the model with labeled target samples $\mathcal{X}_{t}$, which occupy a very small number of the target domain, we fully utilize the majority of unlabeled target samples for regularization.
Recall that the query selection of the proposed ADA framework is based on the loss predictor.
We train the main network (\ie $F$ and $C$) and the auxiliary loss predictor $P$ using unlabeled target samples.
To train the loss predictor $P$, we need ground-truth labels of samples.
Unfortunately, there is no label in the unlabeled target domain $\mathcal{X}_{t_u}$.
To solve this, we utilize target pseudo labels obtained as follows: 
\begin{equation}
\tilde{y}_{t} = \underset{c}{\mathrm{argmax}}(\sigma(C(F(x_{t_u})))),
\label{eq:Pseudo}
\end{equation}
where $\sigma$($\cdot$) is the Softmax function, $c$ is a class index, and $\tilde{y}_{t}$ refers to the target pseudo labels from the model inference.
Note that pseudo labels are updated at pre-determined intervals $\gamma$ instead of every step, and they are used to train the feature extractor $F$ in a self-training manner.
We then use the pseudo-cross-entropy loss to train our loss predictor $P$ using margin ranking loss as in Eq.\ref{eq:predLoss}.

A natural question may arise about the reliability of pseudo labels for model training.
To increase the reliability of the target pseudo labels, we use information maximization considering self-entropy and class diversity, which can be formulated as follows: 
\begin{equation} 
\begin{split}
L_{im}  = 
& -\mathbb{E}_{x_{t_u} \in \mathcal{X}_{t_u}}  \sum_{c=1}^{C}{[\mathbbm{1}_{[c = \hat{y}^t]}  log(\sigma({C(F(x_{t_u}))))}]} \\
& + \xi { D_{KL}(\hat{y_t}, {\frac{1}{C}} \textbf{1}_C) - log(C)}.
\end{split}  
\label{eq:IMLoss}
\end{equation}

The first term represents the self-entropy, which encourages our model to assign disparate one-hot encodings to the feature representations of $\mathcal{X}_{t_u}$.
The second term is used to avoid situations where the target pseudo labels $\tilde{y}_{t}$ are assigned to only a small number of classes, \ie low diversity.
In the above equation,  $\sigma$ denotes Softmax function and $\mathbbm{1}_{[\cdot]}$ is indicator function. $C$ is the number of classes and $\textbf{1}_C$ is a vector with all elements equal to 1 and the same size as the number of classes. 
Importantly, $\hat{y_t} = \mathbb{E}_{x_t\in\mathcal{X}_{t_u}}[\sigma(C(F(x_{t_u})))]$ is the mean output probability of the whole unlabeled target domain.
$D_{KL}$ denotes Kullback–Leibler divergence and $\xi$ stands for controlling variable between two-loss terms.
As a result, our model produces progressively more reliable target pseudo labels $\tilde{y}_{t}$ with our information maximization loss for diverse classes.

To alleviate domain discrepancy between the labeled source domain $\mathcal{X}_{s}$ and unlabeled target domain $\mathcal{X}_{t_u}$, we exploit a conventional mini-max game between the feature extractor $F$ and the domain discriminator $D$. The domain discriminator $D$ attempts to classify the domain label of the given samples while the feature extractor $F$ tries to deceive the domain discriminator $D$ as follows:
\begin{equation}
\begin{split}
L_{dis} = &-\mathbb{E}_{x_s \in \mathcal{X}_{s}}[log D(F(x_s))] \\ 
&-\mathbb{E}_{x_{t_u} \in \mathcal{X}_{t_u}}[log(1 - D(F(x_{t_u})))].
\end{split}
\label{eq:discirmLoss}
\end{equation} 
\begin{equation}
\begin{aligned}
L_{adv} = -L_{dis}.
\end{aligned}
\label{eq:advLoss}
\end{equation}

The total loss consists of the losses mentioned above as:
\begin{equation}
L_{total} = L_{loss} + L_{adv} + L_{im} + L_{dis}.
\label{eq:totalencoder loss}
\end{equation}

As a result, we apply a constraint to generate reliable target pseudo labels while alleviating domain discrepancy.
Importantly, our loss predictor actively utilizes unlabeled target samples occupying the majority of target domains and the whole process is described in Algorithm \ref{alg:stage2}.

\begin{algorithm}[H]
	\SetAlgoLined
	\textbf{Input}:
	labeled source domain: $\mathcal{X}_{s}$, unlabeled target domain: $\mathcal{X}_{t_u}$, trainable modules: ($F$, $P$, $D$), non-trainable module : $C$, total iterations $N$, $\theta_{(F,P,D)}$ : parameters of each module, $\gamma$: intervals for pseudo labeling. \\
	\For{$n=1$ {\bfseries to} $N$}{
		\If{$n \%$  $\gamma$ == 0}{
			get pseudo labels via Eq.\ref{eq:Pseudo}
		}
		 $\hat{d}_i \gets$ $D(F(x_t,x_s))$ \\
		 $\hat{l}_i \gets$ $P(F(x_t))$ \\
		 $l_{G.T} \gets \mathcal{L}_{ce}$ with target pseudo labels \\
		\vspace{3mm}
		 $L_{loss} \gets Ranking  loss$($l_{G.T}$,$\hat{l}_i$) via Eq.\ref{eq:predLoss} \\ 
		 $L_{im} \gets$ $L_{ent}$ + $L_{class\_div}$ via Eq.\ref{eq:IMLoss} \\ 
		 $L_{dis} \gets$ $-log(\hat{d}_i) -log(1 - \hat{d}_i$) via Eq.\ref{eq:discirmLoss} \\
		 $L_{adv} \gets$ $-L_{dis}$ via Eq.\ref{eq:advLoss} \\
		\vspace{3mm}
		  {\bfseries optimize} three modules $F$, $P$, and $D$ in turn: \\
		\vspace{-3mm}
		\begin{small}
			\begin{displaymath}
			\begin{aligned}
			\theta_{F} {\gets} &  \theta_{F} - \lambda \nabla_{\theta_{F}} \big ( \mathcal{L}_{im} + \mathcal{L}_{loss} + \mathcal{L}_{adv}\big)\\ 
			\theta_{P} {\gets} & \theta_{P} - \lambda \nabla_{\theta_{P}} (\mathcal{L}_{loss}) \\
			\theta_{D} {\gets} & \theta_{D} - \lambda \nabla_{\theta_{D}} (\mathcal{L}_{dis}) \\
			\end{aligned}
			\end{displaymath}
		\end{small}
	}
	\caption{Sequential adaptation procedure}  
	\label{alg:stage2}
\end{algorithm}

\section{Experiments}

\vspace{-2mm}
\subsection{Datasets and implementations}
We perform experiments on Digits, Office-31, Office-Home, and VisDA.
\textbf{Digits} mainly consists of two subsets, SVHN (S) \cite{netzer2011reading} and MNIST (M) \cite{lecun1998gradient}. 
SVHN consists of 73,257 RGB images and MNIST consists of 60,000 grayscale images. 
\textbf{Office-31} \cite{saenko2010adapting} consists of 4,652 images with 31 categories collected from three different domains: Amazon (A), Webcam (W), and DSLR (D). 
\textbf{Office-Home} \cite{venkateswara2017deep} consists of 15,588 images collected from four domains with 65 categories: Artistic (A), Clipart (C), Product (P), and Real-World (R).
\textbf{VisDA} \cite{peng2017visda} (2017 Ver.) is a large-scale Sim-to-Real dataset consisting of  280,000 images with 12 categories. 
For a fair comparison with existing methods, we follow the official UDA protocol in all datasets and employ the same architecture for the feature extractor. 
The baseline models for active learning follow the setting of \cite{fu2021transferable}, and we cite the reported results of previous studies if the experimental protocol is the same as ours.

\vspace{-3mm}
\subsection{Comparison results}
\begin{wrapfigure}{h}{0.41\textwidth}
	\vspace{-6mm}
	\centering
	\includegraphics[trim={0 0 0 0},clip, width=0.41\textwidth]{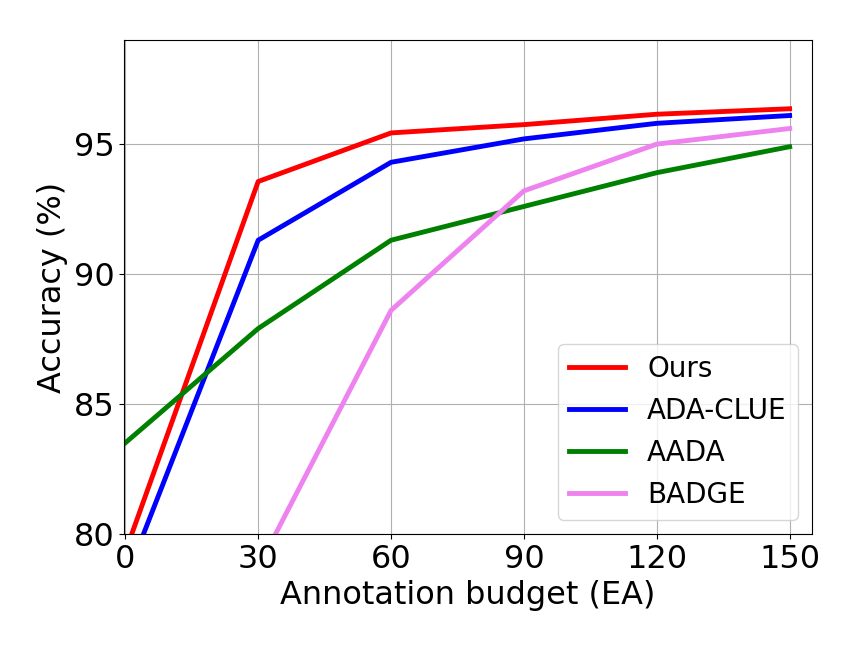}    \vspace{-8mm}
	\caption{   Comparison results on Digits (S $\Rightarrow$ M) with respect to annotation budget.
	}
	\label{fig:s2m}
	\vspace{-1mm}
\end{wrapfigure}
We extensively compare our method with previous ADA methods and various baseline models on four public datasets. 
As shown in  Fig.~\ref{fig:s2m}, our method rapidly increases the performance with very few annotation budgets compared to BADGE and ADA-CLUE on Digits.
AADA has high initial performance, but the growth rate is not steep compared to the proposed method.
From Table~\ref{table:accuracy_officeic}, we can see that the proposed method outperforms various baseline models and the state-of-the-art method (\ie TQS) in Office-31 at 5\% annotation budget. 
Especially, in $A \Rightarrow D$ and $A \Rightarrow W$ scenarios, our method achieves more than 95$\%$ accuracy only using 5$\%$ of annotation budget. 
Given 10$\%$ annotation budget in Office-31, the performance of our method is comparable to TQS as shown in Table~\ref{table:accuracy_officeic_10}.
Also, our proposed method shows state-of-the-art performance in Office-Home and VisDA with 5$\%$ and 10$\%$ annotation budgets significantly as shown in Table~\ref{table:accuracy_officehome} and Table~\ref{table:accuracy_officehome10}.
\begin{wrapfigure}{h}{0.47\textwidth}
	\vspace{-6mm}
	\centering
	\includegraphics[width=0.47
	\textwidth]{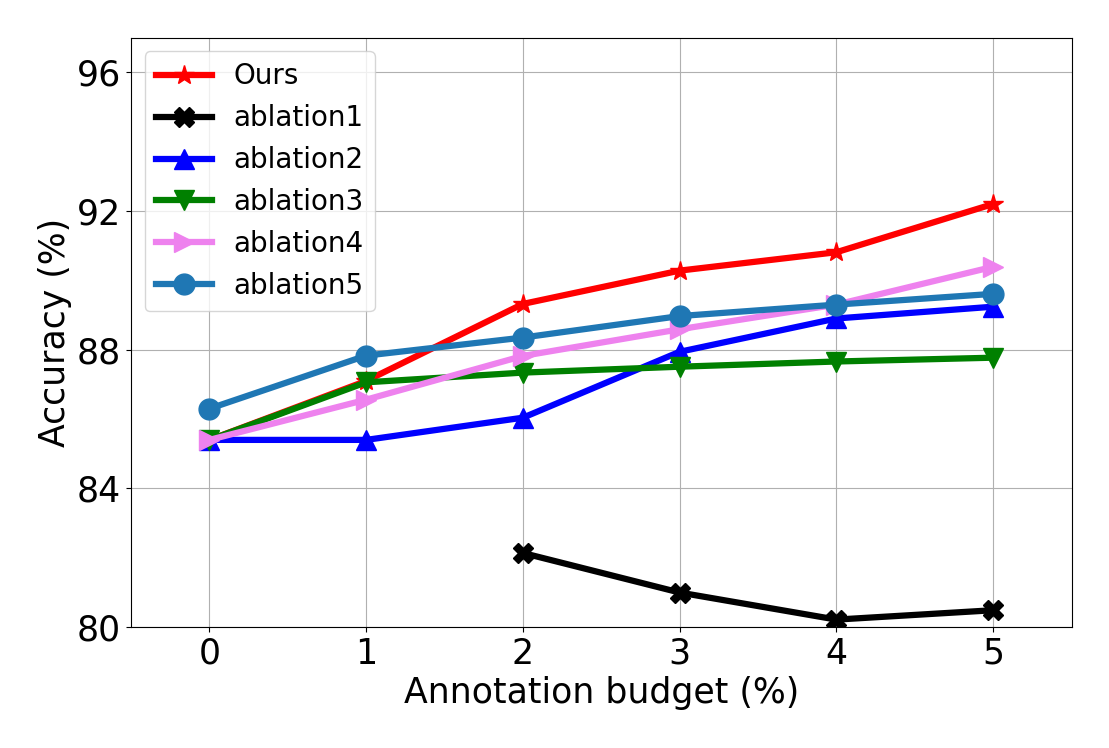}   \vspace{-3mm}
	\caption{   Ablation study on Office-31 with 5$\%$ annotation budget. The setting of each ablation set is presented in Table \ref{table:ablation_settings}. 
	}
	\label{fig:ablation}
	\vspace{-3mm}
\end{wrapfigure}
Note that the performance of S3VAAD was partially reported on the various dataset and therefore not compared all in our experiment. The performance of the ADA-CLUE was not reported in 10$\%$ annotation budget due to the reproducibility issue.

\begin{table*}[]
	\small
	\centering
	\tabcolsep=0.2cm
	\caption{Classification accuracy (\%) on {Office-31} with 5\% annotation budget}
	\label{table:accuracy_officeic}
	\begin{tabular}{lccccccccc}
		\toprule
		\multirow{2}{30pt}{\centering Method}\:\:\:\:\:\:\: &  \multicolumn{7}{c}{Office-31}  \\
		\cmidrule{2-8} 
		& \:A$\Rightarrow$D\: & \:A$\Rightarrow$W\: & \:D$\Rightarrow$A\: & \:D$\Rightarrow$W\: & \:W$\Rightarrow$A\: & \:W$\Rightarrow$D\: & \:Avg\:   \\
		\midrule
		ResNet (source only)~\cite{he2016deep}   & 81.5 & 75.0 & 63.1 & 95.2 & 65.7 & 99.4 & 80.0\\
		\midrule
		RAN (Random Sampling)                    & 87.1  & 84.1 & 75.5 & 98.1 & 75.8  & 99.6 & 86.7\\
		UCN~\cite{joshi2012scalable}             & 89.8  & 87.9 & 78.2 & 99.0 & 78.6  & 100.0 & 88.9\\
		QBC~\cite{dagan1995committee}            & 89.7  & 87.3 & 77.1 & 98.6 & 78.1  & 99.6 & 88.4\\
		Cluster~\cite{nguyen2004active}          & 88.1  & 86.0 & 76.2 & 98.3 & 77.4  & 99.6 & 87.6\\
		Learning loss~\cite{yoo2019learning}     & 80.9  & 84.0 & 69.5 & 98.0 & 69.4 & 99.8 & 83.6 \\
		\midrule
		ADMA~\cite{huang2018cost}                & 90.0 & 88.3 & 79.2 & 100.0 & 79.1 & 100.0 & 89.4\\
		AADA~\cite{su2020active}                 & 89.2 & 87.3 & 78.2 & 99.5 & 78.7 & 100.0 & 88.8\\
		ADA-CLUE~\cite{Prabhu_2021_ICCV}         & 92.0 & 87.3 & 79.0 & 99.2 & 79.6 & 99.8 & 89.5 \\
		S3VAAD~\cite{Rangwani_2021_ICCV}         & 93.0 & 93.7 & 75.9 & 99.4 & 78.2 & 100 & 90.0 \\
		TQS~\cite{fu2021transferable}            & 92.8 & 92.2 & \textbf{80.6} & \textbf{100.0} & 80.4 & \textbf{100.0} & 91.1\\
		\midrule 
		Ours                                   &\textbf{96.6} & \textbf{96.8} & 79.9 & 99.8 & \textbf{81.7} & 99.8 & \textbf{92.2}  \\
		\bottomrule
	\end{tabular}%
	\vspace{-2mm}
\end{table*}

\begin{table*}[]
	\small
	\centering
	\tabcolsep=0.05cm
	\caption{Classification accuracy (\%)  on {Office-Home and VisDA} with 5\% annotation budget}
	\label{table:accuracy_officehome}
	\resizebox{1.0\textwidth}{!}{%
		\begin{tabular}{llccccccccccccc}
			\toprule
			\multirow{2}{30pt}{\centering Method}\:\:\:\:\:\:\: & \multirow{2}{*}{VisDA} & \multicolumn{13}{c}{Office-Home} \\
			\cmidrule{3-15}
			& &{A}$\Rightarrow${C} & {A}$\Rightarrow${P} & {A}$\Rightarrow${R} & {C}$\Rightarrow${A} & {C}$\Rightarrow${P} & {C}$\Rightarrow${R} & {P}$\Rightarrow${A} & {P}$\Rightarrow${C} & {P}$\Rightarrow${R} & {R}$\Rightarrow${A} & {R}$\Rightarrow${C} & {R}$\Rightarrow${P} & \:\:Avg\:\:\\
			\midrule
			ResNet (source only)~\cite{he2016deep}      & 44.7 & 42.1 & 66.3 & 73.3 & 50.7 & 59.0 & 62.6 & 51.9 & 37.9 & 71.2 & 65.2 & 42.6 & 76.6 & 58.3  \\
			\midrule
			RAN (Random Sampling)                       & 78.1 & 52.5 & 74.3 & 77.4 & 56.3 & 69.7 & 68.9 & 57.7 & 50.9 & 75.8 & 70.0 & 54.6 & 81.3 & 65.8 \\
			UCN~\cite{joshi2012scalable}                & 81.3 & 56.3 & 78.6 & 79.3 & 58.1 & 74.0 & 70.9 & 59.5 & 52.6 & 77.2 & 71.2 & 56.4 & 84.5 & 68.2 \\
			QBC~\cite{dagan1995committee}               & 80.5 & 56.9 & 78.0 & 78.4 & 58.5 & 73.3 & 69.6 & 60.2 & 53.3 & 76.1 & 70.3 & 57.1 & 83.1 & 67.9 \\
			Cluster~\cite{nguyen2004active}             & 79.8 & 56.0 & 76.8 & 78.1 & 58.4 & 72.6 & 69.2 & 58.4 & 51.2 & 75.4 & 70.1 & 56.4 & 82.4 & 67.1 \\
			Learning loss~\cite{yoo2019learning}        & 85.8 & 58.2 & 74.2 & 77.4 & 62.6 & 72.8 & 73.4 & 62.1 & 56.6 & 79.6 & 70.7 & 55.1 & 75.9 & 68.1 \\
			\midrule
			ADMA~\cite{huang2018cost}                   & 81.4 & 57.2 & 79.0 & 79.4 & 58.2 & 74.0 & 71.1 & 60.2 & 52.2 & 77.6 & 71.0 & 57.5 & 85.4 & 68.6 \\
			AADA~\cite{su2020active}                    & 80.8 & 56.6 & 78.1 & 79.0 & 58.5 & 73.7 & 71.0 & 60.1 & 53.1 & 77.0 & 70.6 & 57.0 & 84.5 & 68.3 \\
			ADA-CLUE~\cite{Prabhu_2021_ICCV}            & 85.2 & 63.6 & 79.3 & 80.9 & 68.8 & 77.5 & 76.7 & 66.3 & 57.9 & 81.4 & 75.6 & 60.8 & 86.3 & 72.5  \\
			S3VAAD~\cite{Rangwani_2021_ICCV}            & 77.7 & 57.3  & 73.9 & 76.6 & 60.3 & 76.5 & 71.1 & 57.6 & 56.0 & 78.7 & 71.4 & 63.1 & 83.3 & 68.8  \\ 
			TQS~\cite{fu2021transferable}               & 83.1 & 58.6 & 81.1 & 81.5 & 61.1 & 76.1 & 73.3 & 61.2 & 54.7 & 79.7 & 73.4 & 58.9 &  \textbf{86.1} & 70.5 \\
			\midrule 
			Ours                                         & \textbf{86.8} & \textbf{63.7} & \textbf{83.9} & \textbf{82.5} & \textbf{69.7} & \textbf{82.7} & \textbf{81.4} & \textbf{70.3} & \textbf{61.2} & \textbf{84.6} & \textbf{77.4} & \textbf{63.4} & 85.9 & \textbf{75.6}\\
			\bottomrule
	\end{tabular}}
\end{table*}

\begin{table*}[t]
	\small
	\centering
	\tabcolsep=0.2cm
	\caption{Classification accuracy (\%) on {Office-31} with 10\% annotation budget}
	\label{table:accuracy_officeic_10}
	\begin{tabular}{lccccccccc}
		\toprule
		\multirow{2}{30pt}{\centering Method}\:\:\:\:\:\:\: &  \multicolumn{7}{c}{Office-31}  \\
		\cmidrule{2-8} 
		& \:A$\Rightarrow$D\: & \:A$\Rightarrow$W\: & \:D$\Rightarrow$A\: & \:D$\Rightarrow$W\: & \:W$\Rightarrow$A\: & \:W$\Rightarrow$D\: & \:Avg\:   \\
		\midrule
		ADMA~\cite{huang2018cost}                & 94.0 & 93.4 & 84.4 & 100.0 & 84.6  & 100.0 & 92.7\\
		AADA~\cite{su2020active}                 & 93.5 & 93.1 & 83.2 & 99.7  & 84.2  & 100.0 & 92.3\\
		ADA-CLUE~\cite{su2020active}                 & 93.5 & 93.1 & 83.2 & 99.7  & 84.2  & 100.0 & 92.3\\
		S3VAAD~\cite{Rangwani_2021_ICCV}         & \textbf{98.0} & 95.6 & 81.0 & 99.4 & 81.6     & 100.0     & 92.6   \\
		TQS~\cite{fu2021transferable}            & 96.4 & 96.4 & \textbf{86.4} & \textbf{100.0} & \textbf{87.1}  & 100.0 & \textbf{94.4}\\
		\midrule 
		Ours                                   & 97.8 & \textbf{97.9} & 85.0 & 99.8 & 85.3 & \textbf{100.0} & 94.3  \\
		\bottomrule
	\end{tabular}%
\end{table*}

\begin{table*}[]
	\small
	\centering
	\tabcolsep=0.05cm
	\caption{Classification accuracy (\%)  on {Office-Home and VisDA} with 10\% annotation budget}
	\label{table:accuracy_officehome10}
	\resizebox{1\textwidth}{!}{%
		\begin{tabular}{llccccccccccccc}
			\toprule
			\multirow{2}{30pt}{\centering Method}\:\:\:\:\:\:\: & \multirow{2}{*}{VisDA} & \multicolumn{13}{c}{Office-Home} \\
			\cmidrule{3-15}
			& &{A}$\Rightarrow${C} & {A}$\Rightarrow${P} & {A}$\Rightarrow${R} & {C}$\Rightarrow${A} & {C}$\Rightarrow${P} & {C}$\Rightarrow${R} & {P}$\Rightarrow${A} & {P}$\Rightarrow${C} & {P}$\Rightarrow${R} & {R}$\Rightarrow${A} & {R}$\Rightarrow${C} & {R}$\Rightarrow${P} & \:\:Avg\:\:\\
			\midrule
			ADMA~\cite{huang2018cost}                       & 84.8 & 66.5 & 85.4 & 82.8 & 63.8 & 80.9 & 76.3 & 67.7 & 61.6 & 80.9 & 74.3 & 66.8 & 89.7 & 74.7\\
			AADA~\cite{su2020active}                        & 84.6 & 65.8 & 84.5 & 82.2 & 64.1 & 80.6 & 76.1 & 67.6 & 62.6 & 80.1 & 73.7 & 66.1 & 88.6 & 74.3\\
			S3VAAD~\cite{Rangwani_2021_ICCV}                &  81.1  & 64.6    & 81.4    & 80.6    & 62.6 & 82.8    & 76.2    & 61.7    & 62.2    & 81.9    & 73.0    & 65.3    & 87.1    & 73.3  \\ 
			TQS~\cite{fu2021transferable}                   & 87.2 & 68.0          & 87.7          & 85.7          & 67.0          & 83.0          & 78.7          & 69.3          & 64.5          & 83.9         & 77.8 & 68.9 & \textbf{90.6} & 77.1\\
			\midrule 
			Ours                                            & \textbf{90.2} & \textbf{70.7} & \textbf{87.9} & \textbf{86.9} & \textbf{74.3} & \textbf{87.4} & \textbf{85.4} & \textbf{74.5} & \textbf{69.2} & \textbf{87.4} & \textbf{81.4} & \textbf{70.2} & 90.4 & \textbf{80.5} \\
			\bottomrule
		\end{tabular}%
	}
\end{table*}

\begin{figure*}[t]
	\label{fig:diversity}
	\begin{center}
		\def\arraystretch{0}
		\resizebox{1\textwidth}{!}{%
			\begin{tabular}{@{}c@{}c@{}c@{}c@{}c}
				\includegraphics[width=0.315\linewidth]{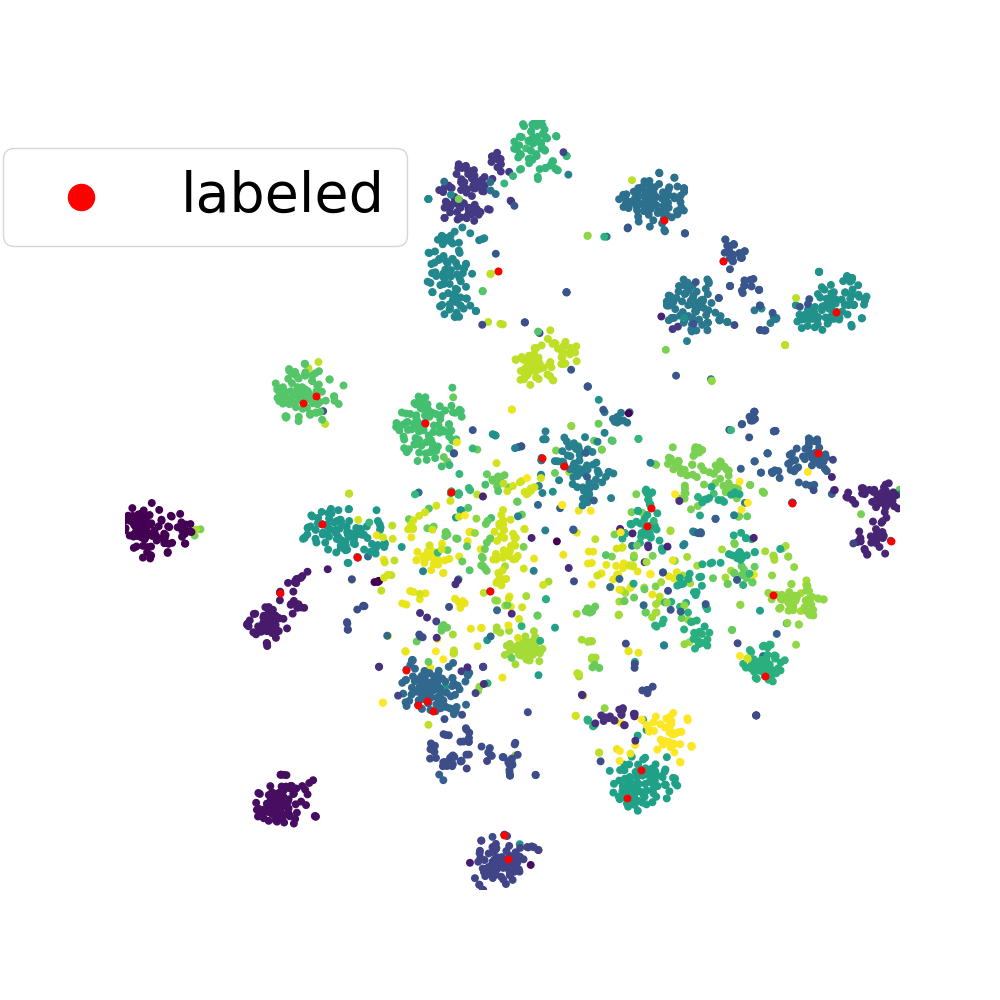}&
				\includegraphics[width=0.315\linewidth]{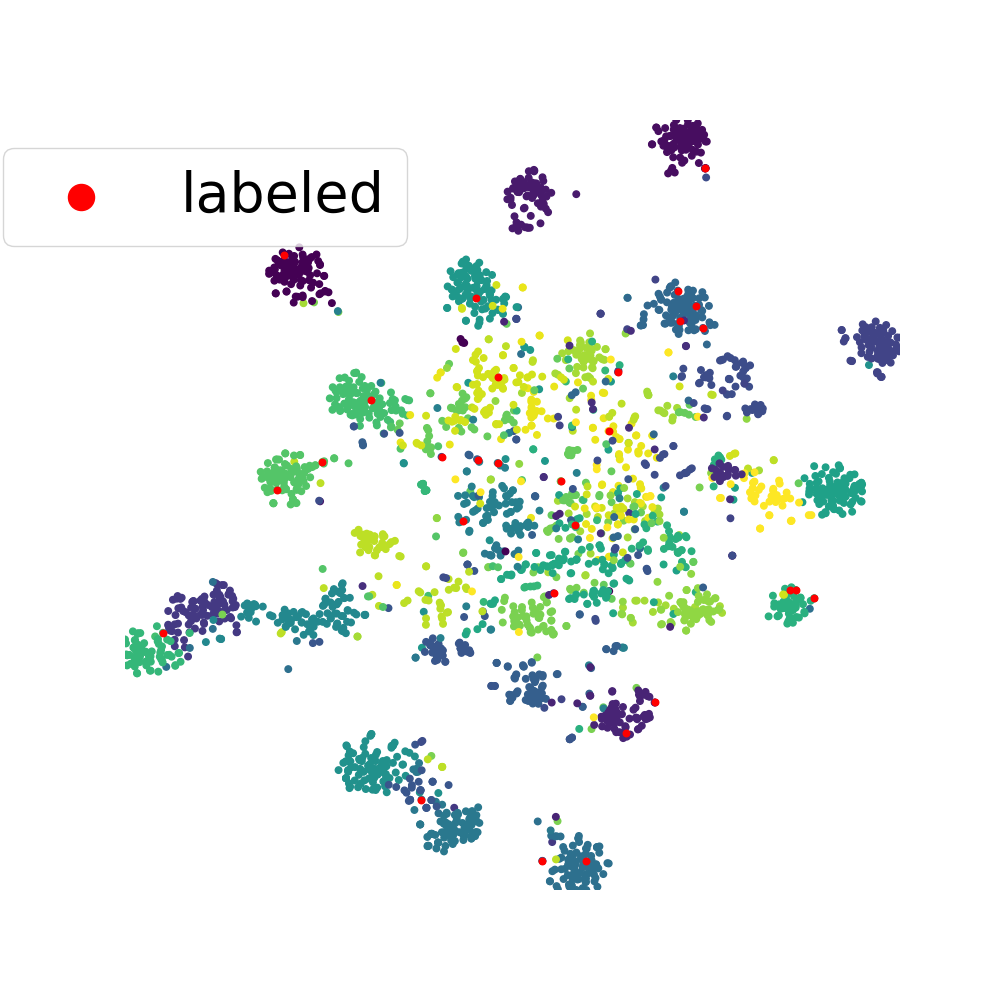} &
				\includegraphics[width=0.315\linewidth]{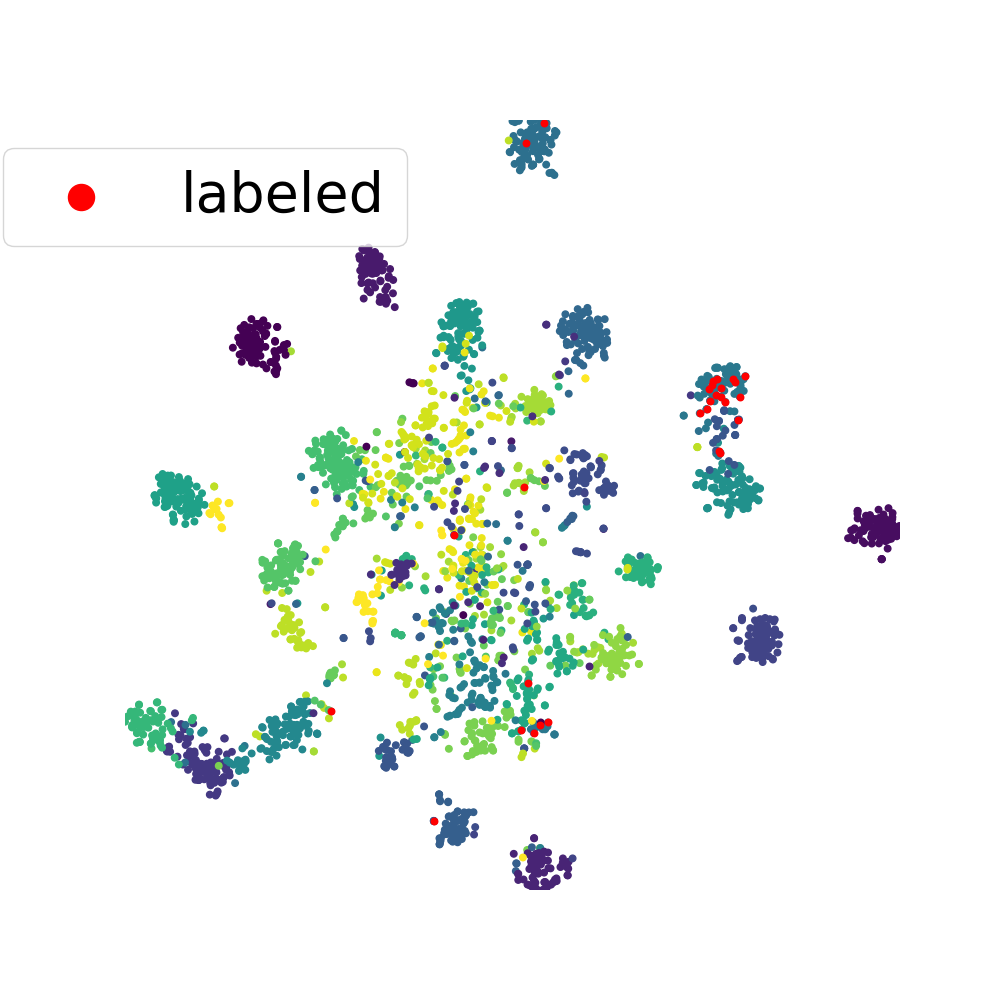} &
				\includegraphics[width=0.315\linewidth]{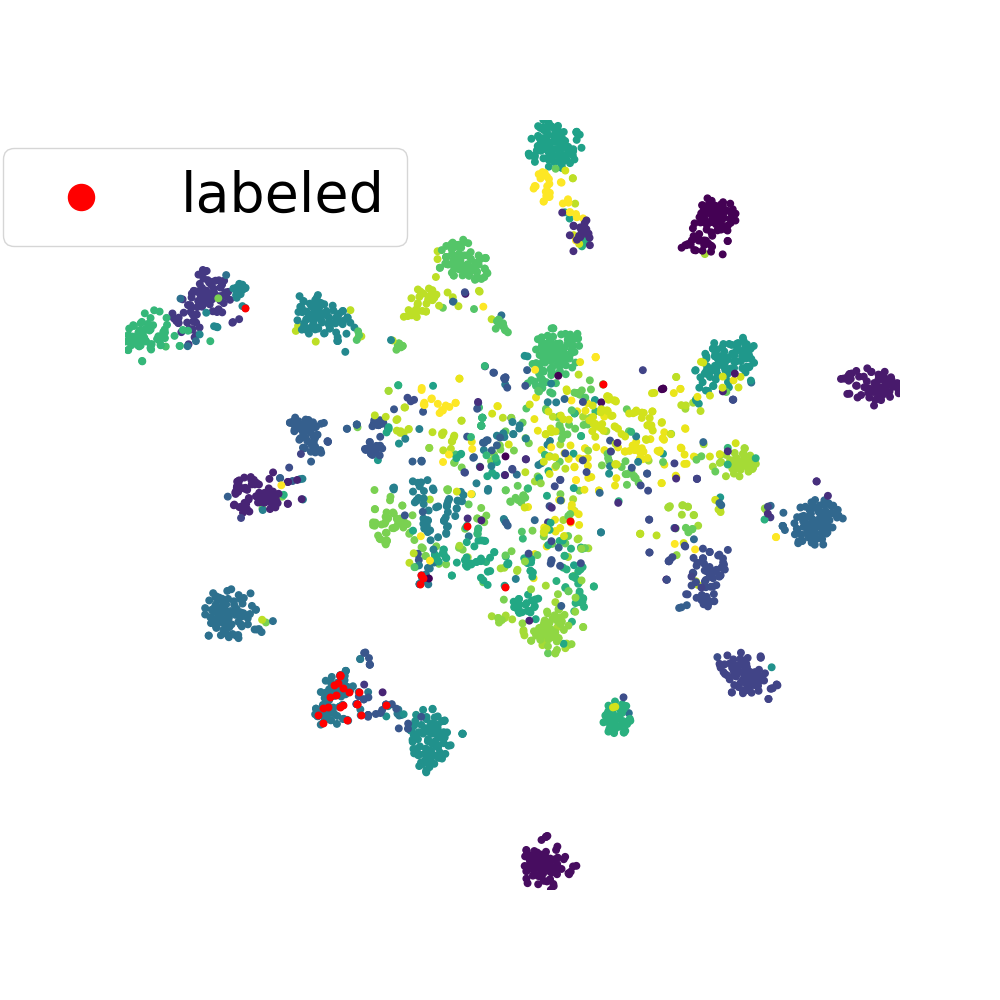} & \\
				{(a) Random selection} & {(b) Random selection } & {(c) learning loss 
					\cite{yoo2019learning} } & {(d) learning loss \cite{yoo2019learning}}
			\end{tabular}
		}
	\end{center}
	\caption
	{
		t-SNE visualization on Office-31 ({D}$\Rightarrow${A}). (a) and (b) visualize the result of random query selection while (c) and (d) is the visualized results using plain loss-based query selection. The red dots represent labeled target samples by query selection, and the colors indicate each class.
	}
	\label{fig:tsne}
	\vspace{-4mm}
\end{figure*}

\begin{table}[]
	\noindent\begin{minipage}{\textwidth}
		\begin{minipage}{0.48\textwidth}
			\centering
			\caption{Ablation study setting ({RAN : random selection, L : loss-based selection, S$_1$ : selected sample training, S$_2$-L : sequential adaptation w/ [G.T label or pseudo label], S$_{2}$-IM : sequential adaptation w/ IM loss}).}
			\vspace{-0.5mm}
			\resizebox{0.95\textwidth}{!}{%
				\begin{tabular}{@{}ccccc@{}}
					\toprule
					& Query selection & S$_1$  & S$_2$-L & S$_{2}$-IM \\ \midrule
					\multicolumn{1}{c} {Ablation1} & RAN + L & X & G.T  & X \\
					\multicolumn{1}{c} { Ablation2}  & RAN + L & O & G.T & X \\
					\midrule
					\multicolumn{1}{c} { Ablation3} & L & O &  G.T & X \\
					\multicolumn{1}{c} {Ablation4} & L & X &  P.L & O \\
					\multicolumn{1}{c} { Ablation5} & L & swap-S$_2$ & swap-S$_1$ & X \\
					\midrule
					\multicolumn{1}{c} {Ours} & L & O &  O & O \\ \bottomrule
				\end{tabular}%
			}
			\label{table:ablation_settings}
		\end{minipage}
				\hfill
			\begin{minipage}{0.48\textwidth}
			\centering
			\caption{Accuracy (\%) change with respect to annotation budget $B$\% and pseudo label update intervals $\gamma$ in Office-31. }
			\vspace{2mm}
			\resizebox{0.95\textwidth}{!}{%
				\begin{tabular}{lcccccc}
					\toprule
					& $\gamma$ = 2 & $\gamma$ = 10 & $\gamma$ = 20 & $\gamma$ = 30   \\ \midrule
					\multicolumn{1}{c|}{$B$ = 0} & 87.6       & \textbf{88.6} & 87.7                & 87.7       \\
					\multicolumn{1}{c|}{$B$ = 1} & 91.0       & 91.8          & \textbf{91.8}       & 91.0         \\
					\multicolumn{1}{c|}{$B$ = 2} & 93.4       & 92.2          & \textbf{94.8}       & 91.8       \\
					\multicolumn{1}{c|}{$B$ = 3} & 94.6       & 93.6          & \textbf{95.6}       & 93.2         \\
					\multicolumn{1}{c|}{$B$ = 4} & 95.0       & 94.2          & \textbf{95.6}       & 93.4         \\
					\multicolumn{1}{c|}{$B$ = 5} & 95.0       & 95.0          & \textbf{96.0}       & 94.2         \\ \bottomrule
				\end{tabular}
			}
			\label{table:pseudo-label}
		\end{minipage}
	\end{minipage}
	\vspace{-4mm}
\end{table}

\vspace{-2mm}
\subsection{Ablation study and analysis}
\label{ablation}

To validate each module of the proposed method, we evaluate diverse experimental settings on Office-31 as shown in Fig. \ref{fig:ablation}. 
Firstly, we set the experiment as loss-based active learning (AL) schemes, which start with Cold-Start (\ie random selection at the initial step) and train the model with only labeled data.  
The second ablation is corresponding to Cold-Start with {S$_1$ and S$_2$ without utilizing pseudo label and IM loss training.} By comparing Ablation1 and Ablation2, we found that sequential learning is effective for active domain adaptation.   
The result of Ablation3 shows the performance degradation when training the model only with selected high-loss samples.
Ablation4 supports the effectiveness of using pseudo labels from the majority of unlabeled target samples. 
The last ablation shows that the proposed sequential order has the best performance between S$_1$ to S$_2$ and S$_2$ to S$_1$ settings. Note that S$_1$ conducts domain adaptation and pseudo labeling process like algorithm \ref{alg:stage2} and S$_2$ instead conducts selected sampling training in Ablation4.
Overall, we empirically demonstrate the effectiveness of our sequential learning and reliable pseudo labels by information loss.

\subsection{Visualization of sample diversity}
\label{ssec:diversity_issue}
We claim that loss-based query selection cannot handle the sample diversity issues in domain shift scenarios. To support this, we visualize selected target samples in the target domain according to random selection and loss-based query selection. 
From Fig. \ref{fig:tsne}, we can see that the samples selected by loss-based query selection are clustered in some regions rather than spread out in the target domain compared to the randomly selected samples.
Although sample diversity is low when loss-based query selection is used alone, diversity can be increased by using the proposed model regularization, which leads to improved performance.

\vspace{-2mm}

\subsection{Parameter sensitivity analysis}
\label{sssec:parameter_sensitivity}

\begin{figure}[t]
	\begin{center}
		\def\arraystretch{0.5}
		\begin{tabular}{@{}c@{}c@{}c}
			\hspace{-5mm}
			\includegraphics[width=0.315\linewidth]{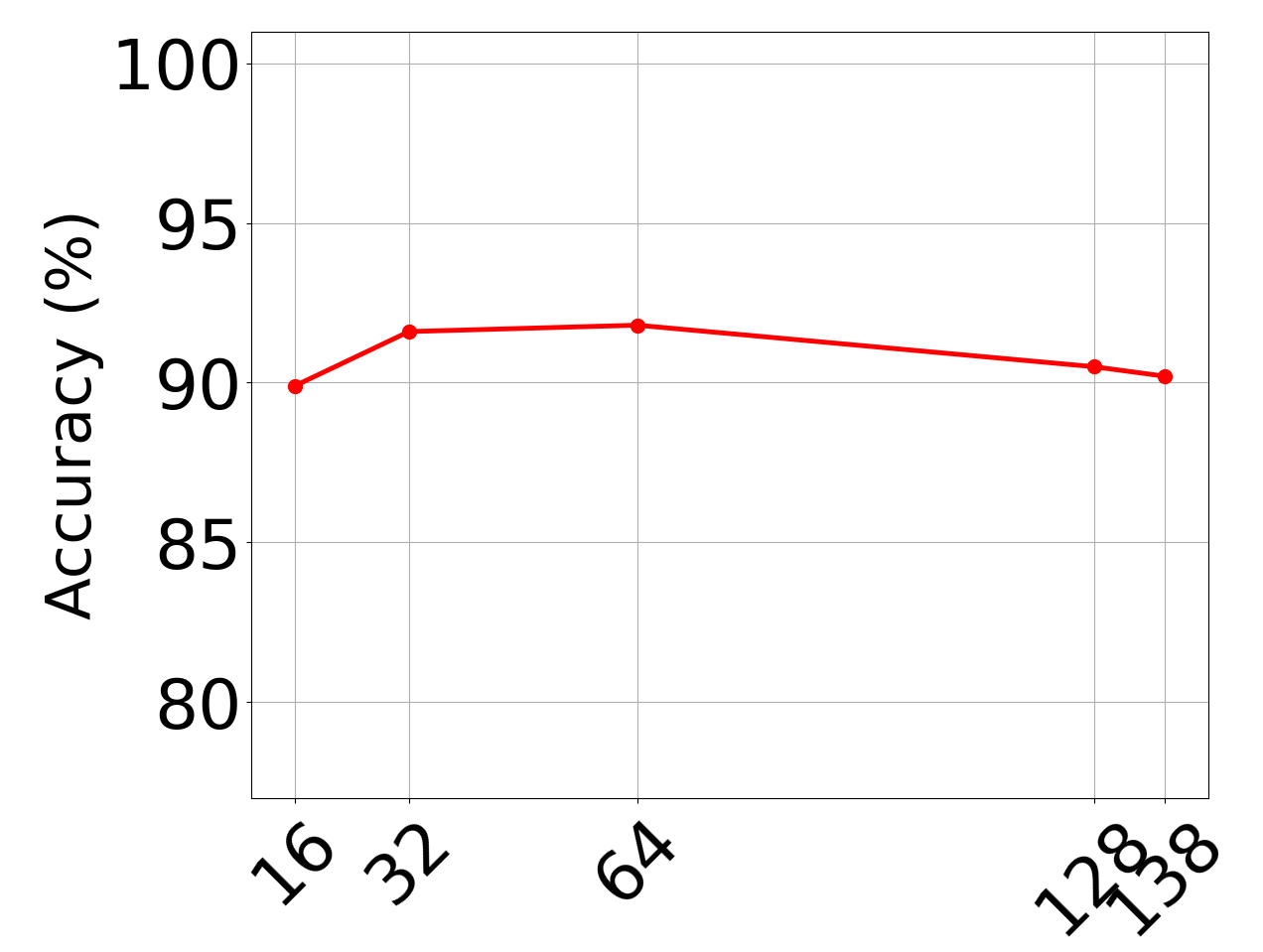} &
			\includegraphics[width=0.315\linewidth]{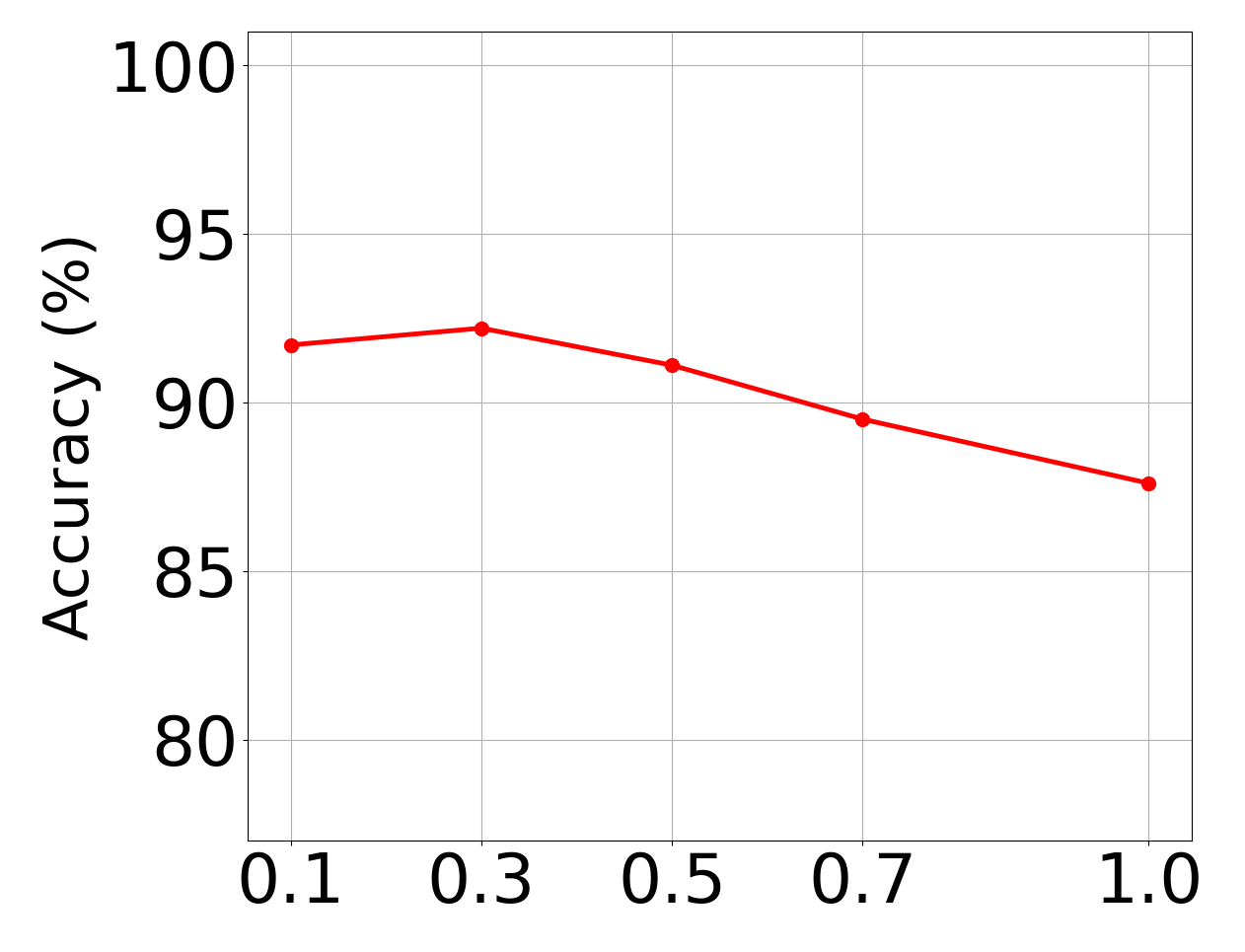} &
			\includegraphics[width=0.315\linewidth]{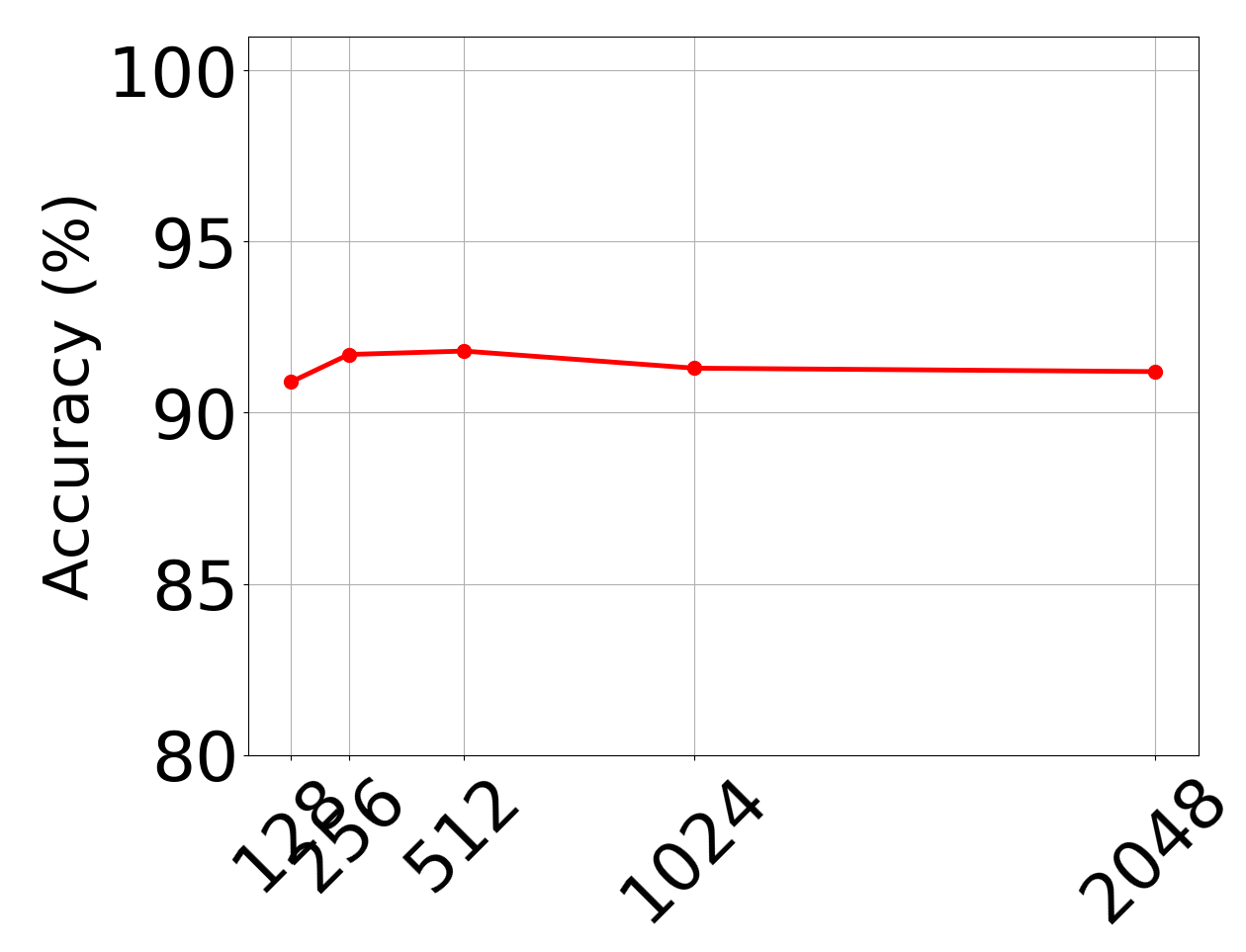}
			\\
			\vspace{-2mm}
			{(a)}  & {(b) } & {(c)}\\
		\end{tabular}
		\vspace*{-0.4mm}
	\end{center}
	\caption{ Parameter sensitivity analysis on Office-31. (a) batch size, (b)  balancing parameter $\xi$, (c) loss predictor's  dimension. }
	\label{fig:three figures}
\end{figure}

We present sensitivity analysis on hyperparameters including batch size, balance parameter $\xi$ for information maximization loss, and intermediate dimension for the loss predictor.
Fig. \ref{fig:three figures} implies that the batch size or intermediate dimension for the loss predictor is not sensitive, and low self-entropy is more important in information maximization loss.
Furthermore, we analyze pseudo-label-update interval parameter $\gamma$.
From Table~\ref{table:pseudo-label}, we can confirm that updating the pseudo label too often slows down the learning time, but does not improve the performance.

\subsection{Pseudo-label reliability}
\label{sssec:pseudo_label_reliability}
\begin{wrapfigure}{h}{0.47\textwidth}
	\vspace{-9mm}
	\centering
	\includegraphics[width=0.465
	\textwidth]{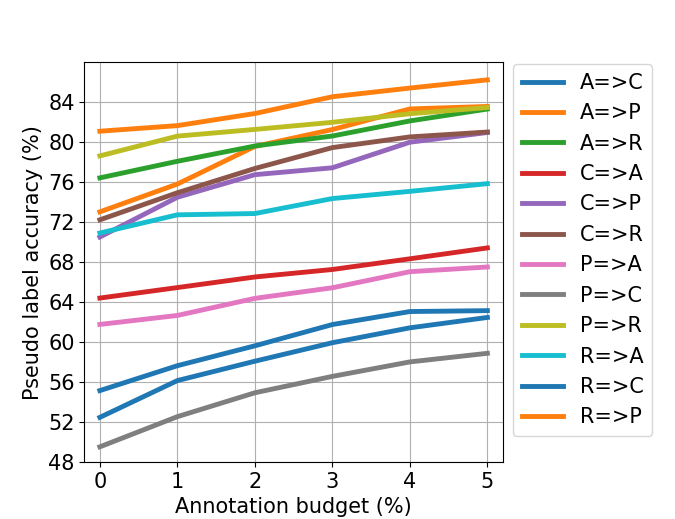}
	\vspace{-3mm}
	\caption{    Pseudo labels accuracy ($\%$) in Office-Home with 5$\%$ of annotation budget. 
	}
	\label{fig:last_ablation}
\end{wrapfigure}
We actively use target pseudo labels as the ground truth label for loss prediction and adaptation. To increase the reliability of pseudo labels, we encourage pseudo labels to have low self-entropy and diverse class distributions.
As shown in Fig. \ref{fig:last_ablation}, our proposed method gradually improves the reliability of the pseudo label for the unlabeled target domain across all the adaptation scenarios in Office-Home.
Surprisingly, half of the total 12 adaptation scenarios show an accuracy of over 80\% when the annotation budget is 5\%.
The result of increasing pseudo-label accuracy as the annotation budget increases demonstrates the effectiveness of the proposed ADA framework including information loss with sample diversity.
\vspace{-2mm}

\vspace{-2mm}
\section{Conclusion} 
While previous active domain adaptation methods focus on sample selection criteria, \eg uncertainty, diversity, and committee, we exploit loss-based query selection and propose model regularization schemes.
The main difference between the proposed method and individual-related work is the sequential learning scheme considering domain type (source/target) and labelness (labeled/unlabeled).
We first train our main network including an auxiliary loss predictor with a small number of ground-truth labels by oracle. We then fully utilize numerous pseudo labels where the reliability is improved by information maximization loss.
We extensively show the limitations of applying only loss-based query selection to active domain adaptation and extensively present analysis for our method.
Our model achieves state-of-the-art performance in various active domain adaptation scenarios.

\bibliographystyle{splncs}
\bibliography{ref}
\end{document}